\documentclass[conference]{IEEEtran}
\IEEEoverridecommandlockouts
\usepackage{cite}
\usepackage{amsmath,amssymb,amsfonts}
\usepackage{algorithmic}
\usepackage{graphicx}
\usepackage{textcomp}
\usepackage{xcolor}
\usepackage{verbatim}
\usepackage{soul}
\usepackage{multirow}
\def\BibTeX{{\rm B\kern-.05em{\sc i\kern-.025em b}\kern-.08em
    T\kern-.1667em\lower.7ex\hbox{E}\kern-.125emX}}

\begin{document}

\title{Neural Medication Extraction: A Comparison of Recent Models in Supervised and Semi-supervised Learning Settings
\thanks{This work has been partially supported by MIAI@Grenoble Alpes (ANR-19-P3IA-0003) funded by the French program Investissement d'avenir.}
}
\author{\IEEEauthorblockN{Ali Can Kocabiyikoglu,\\Jean-Marc Babouchkine}
\IEEEauthorblockA{Calystene SA, 38320 Eybens, France \\
a.kocabiyikoglu@calystene.com,\\
jm.babouchkine@calystene.com}
\and
\IEEEauthorblockN{François Portet}
\IEEEauthorblockA{Univ. Grenoble Alpes, CNRS, Grenoble INP\\ LIG F-38000 Grenoble France\\
francois.portet@imag.fr}
\and
\IEEEauthorblockN{Raheel Qader}
\IEEEauthorblockA{Lingua Custodia, Paris, France \\
raheel.qader@gmail.com}
}

\maketitle

\begin{abstract}
Drug prescriptions are essential information that must be encoded in electronic medical records. However, much of this information is hidden within free-text reports. This is why the medication extraction task has emerged. To date, most of the research effort has focused on small amount of data and has only recently considered deep learning methods. In this paper, we present an independent and comprehensive evaluation of state-of-the-art neural architectures on the I2B2 medical prescription extraction task both in the supervised and semi-supervised settings. The study shows the very competitive performance of simple DNN models on the task as well as the high interest of pre-trained models. Adapting the latter models on the I2B2 dataset enables to push medication extraction performances above the state-of-the-art. Finally, the study also confirms that semi-supervised techniques are promising to leverage large amounts of unlabeled data in particular in low resource setting when labeled data is too costly to acquire.
\end{abstract}


\section{Introduction}\label{sec:intro}

\begin{figure}[htbp]
    \centering
    \frame{\includegraphics[width=8.8cm]{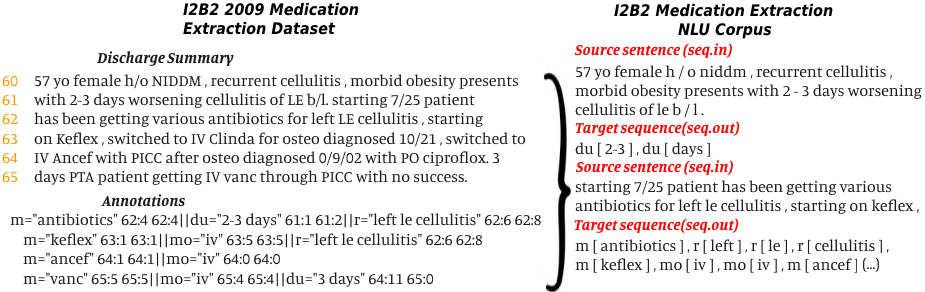}}
    \caption{Left: an I2B2 discharge summary excerpt and its medication annotations. Right: same excerpt segmented by sentence and formatted for sequence-to-sequence processing.
(\tiny{m = name; do = dose; mo = mode of administration; f = frequency; du = duration; r= reason})}
    \vspace{-0.3cm}
    \label{fig:annot}
\end{figure}

In electronic health records (EHR) and other medical documents, drug information is often recorded in clinical notes, making it difficult for computerized applications to access this information as part of daily health care. Automatically extracting structured information related to drug prescriptions from medical free-texts is known as the medication extraction task. This task has attracted attention from the NLP community since the emergence of the I2B2 2009 medication extraction challenge~\cite{uzuner2010community}. In this challenge, the goal was to automatically label chunks of medication information from a whole clinical document. Figure \ref{fig:annot} (left) shows an excerpt of a discharge summary in I2B2 from which information such as prescription duration and medication name should be extracted. Such kind of medication extraction system could be very useful to medical prescription writing software that are used to reduce the number of errors during the prescription, the transcription and the administration process of drugs.

Since the I2B2 2009 challenge, most work has used rule-based systems with the exception of a few hybrid ones. Some recent approaches using deep neural networks have shown promising results~\cite{dai2017medication,fan2020adverse,si2019enhancing}. However, the task lacks a comprehensive comparison of deep learning methods over more traditional methods. Furthermore, the data provided within the I2B2 challenge is still very small for the needs of deep model training. Hence, methods able to leverage large amount of unlabeled clinical data (e.g., MIMIC III~\cite{johnson2016mimic}) should be evaluated on this task. Among these methods, the recent pre-trained models have not been systematically studied for the task. Furthermore, recent semi-supervised techniques have also been rarely applied \cite{tao2018fable,qader2019semi}.

In this paper, we explore the benefit of pre-trained models and semi-supervised learning to leverage non-annotated clinical documents for deep learning models. Indeed, for such a task, data annotation process requires medical experts and often patient data that has to be anonymised. This makes the process costly and limits the distribution of datasets. 
This is why methods resistant to OOV words such as BPE (Byte Pair Encoding) must be evaluated. Indeed, even in such a narrow drug prescription domain, vocabulary size and rare words for the input sequence are numerous~\cite{zhang2019biowordvec}. This is why methods resistant to OOV words such as BPE (Byte Pair Encoding) must be evaluated.

\textbf{Paper contributions.}
Our objective is to provide a comprehensive evaluation of the standard deep learning seq2seq methods (including Transformers) on the I2B2 2009 medication extraction task. We include a focused evaluation of pre-trained models and semi-supervised training to leverage unannotated medical corpora. The results show performances above the state-of-the-art for pre-trained models and competitive performances of semi-supervised training.

\textbf{Outline.}
The remaining of the paper presents a short review of the state-of-the-art before describing the methodology to pre-process the corpora and to perform supervised and semi-supervised learning. The experiment section then shows that the BlueBert pre-trained model 
can reach performance beyond the current state of the art. We then finish the paper by a short discussion and presentation of future work.

\section{Related Work}\label{sec:sota}

The only known benchmarking efforts in medication extraction are the I2B2 2009 challenge~\cite{uzuner2010community} along with n2c2 shared task~\cite{henry20202018}. It is only recently that 
deep neural networks have superseded rule-based or hybrids models. 
Most significant progress has been made both thanks to deep models and also to the ability to leverage larger amount of data either through pre-trained embedding or through semi-supervised training.  Regarding the pre-trained models, word embeddings trained on the MIMIC-III database have been used to improve slightly the state-of-the-art on the I2B2 medication extraction task~\cite{tao2017prescription}. The use of more performing pre-trained models, such as ELMo or BERT~\cite{peters2018deep,devlin2018bert} for word embeddings, have become prevalent in the biomedical NLP domain exceeding benchmarks on certain tasks~\cite{lee2020biobert,fan2020adverse,mulyar2020mt}. For instance, Yang \emph{et al.} \cite{Yang2020}, have compared 4 transformers models BERT, ALBERT, RoBERTa, and ELECTRA on a relation extraction task and showed the definite superiority of those. In a similar vein, Peng \emph{et al.}~\cite{peng2019transfer} have proposed a benchmark setting (called BLUE) to evaluate pre-trained model in a clinical setting, showing that BERT model pre-trained on PubMed abstracts and MIMIC-III was superior (we refer to it as BlueBert in the rest of the paper). 
Recently, biomedical contextual embeddings have also been applied to improve the performance of adverse drug identification and the medication extraction task~\cite{narayanan2020evaluation,henry20202018}.

From a semi-supervised point of view, Tao \emph{et al.}~\cite{tao2018fable} have proposed a semi-supervised system that achieved the current best overall performance on the I2B2 2009 medication extraction task by leveraging the non-annotated part of corpus (and also by using human annotations). More recently,  Guzman \emph{et al.}~\cite{guzman2020assessment} proposed a system based on a LSTM model and transfer learning claiming state-of-the-art performance on extracting specific entities of the I2B2 dataset. However, their system is only partially sketched and thus difficult to reproduce. On other tasks, such as biomedical relation extraction, variational autoencoders~\cite{zhang2019exploring} or event feature coupling generalization (EFCG)~\cite{wang2013semi} have been proposed to benefit from unannotated datasets. In particular, \cite{amin2020data} show that for the biomedical relation extraction task, a distantly supervised approach enables to produce large amounts of labeled but noisy data can be leveraged efficiently for data-driven approach. Despite, these recent progress, semi-supervised learning for medication extraction has only been applied by Tao \emph{et al.} \cite{tao2018fable}, which is to-date the state-of-the-art.

\begin{figure*}[!htb]
    \centering
    \includegraphics[width=0.75\textwidth]{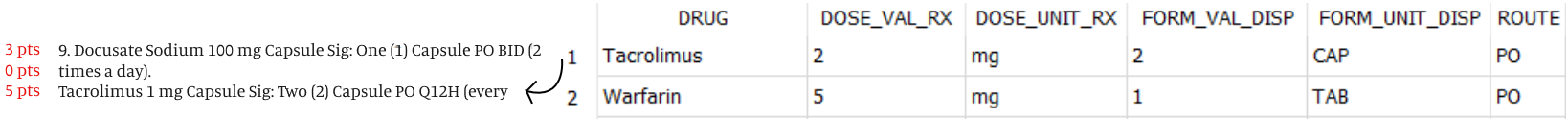}
    \caption{Example matching of medication records database and sentences from MIMIC-III clinical texts}
    \vspace{-0.2cm}
    \label{fig:mimicData}
\end{figure*}
From this brief related work section, it is clear that deep neural network methods have shown significant progress in several bioNLP tasks. However, the ability to leverage larger amount of data has not been fully explored on the I2B2-2009 medication extraction task using recent methods. This paper is an attempt to not only propose a model taking advantage of the most recent NLP advances but also to provide comprehensive evaluation of state-of-the-art models on this task.

\section{Method}

The medication extraction problem is often addressed as a sequence labeling task requiring aligned BIO format. This is extremely costly to annotate and it prevents models from abstracting since they must stick to the surface form of the textual input. 
In this work, we approach the problem of medication extraction through encoder-decoder seq2seq models where input text is first abstracted by the encoder and then the labels and values are generated by the decoder. Figure~\ref{fig:annot} shows on the left side an excerpt of an I2B2 discharge summary. In that case the annotations consist of slot values indexed by position within the document (line:character index). The right side of Figure~\ref{fig:annot} shows the result of the I2B2 sentences adaptation. The document was sentence segmented and for each sentence the annotation was unaligned. This format is more realistic with respect what can be found in real clinical data were free-text segments might be only loosely related to electronic records. 

\subsection{Corpora and data pre-processing}

The medication extraction models were trained on two publicly available datasets: the I2B2 2009 medication extraction dataset and the MIMIC-III dataset.

\subsubsection{I2B2 dataset}

The I2B2 dataset is composed of 1243 clinical documents. The annotation guidelines established for the I2B2 challenge focused on identifying the name of medication, their dosage, their mode of administration, frequency, duration and reason for administration in discharge summaries~\cite{uzuner2010community}. In the official distribution, 10 documents were annotated by medical experts and 251 by the research community. These latter 251 documents became the official test corpus for the evaluation~\cite{uzuner2010community}. The remaining 982 documents were un-annotated.  

\subsubsection{MIMIC-III unannotated dataset}

MIMIC-III~\cite{johnson2016mimic} is a large clinical dataset that does not have annotation data fit for supervised machine learning. But it does contain medication information both in the textual part of the dataset and in the database part (medication records). For the need of the semi-supervised learning, this corpus 
required some pre-processing to extract the relevant pieces of information related to drug prescriptions both from the textual point of view and the database point of view. To do so, we set up a simple algorithm: For each prescription in the database, the lines of the discharge summary of the same a patient's id were scored using regular expressions. Take the example Figure \ref{fig:mimicData}. The right side shows an extract of the prescription table while the left side shows some sentences of the patient's discharge summary that were scored. The line with 5pts had matched the most features of the Tacrolimus line. Thus, both the sentence and the prescription line are added to the annotated dataset. Prescriptions not matching any line of the discharge summary are not added to the unannotated dataset.

\begin{table}[!tbh]
\centering
{\tiny
\begin{tabular}{|c|c|c|c|c|c|}
\hline
\multirow{3}{*}{Data Split} & \multicolumn{4}{c|}{I2B2-2019} & MIMIC-III \\ \cline{2-6} 
& \multicolumn{3}{c|}{supervised} & semi-supervised & semi-supervised \\
\cline{2-6} 
  & \begin{tabular}[c]{@{}c@{}}community\\ annotations\end{tabular} & \begin{tabular}[c]{@{}c@{}}expert\\ annotations\end{tabular} & \begin{tabular}[c]{@{}c@{}}unannotated\\ set\end{tabular} & \begin{tabular}[c]{@{}c@{}}held out\\ unannotated set\end{tabular} & clinical notes \\

 \hline
train & - & - & 22,907 sents  & 4,655 sents & 257,811 sents \\ \hline
validation & & 148 sents & 2,158 sents & -  & 2,605 sents \\ \hline
test & 4411 sents &  &   & - & - \\ \hline
\end{tabular}
}
\vspace{0.1cm}
\caption{Final distribution of the corpora for supervised and unsupervised learning.}
\vspace{-0.5cm}
\label{tab:dataDistributions}
\end{table}

At the end of the process, 962,252 lines of sentences related to the database records were extracted. It was then further filtered to remove similar examples as well as too long ones 
to obtain a final corpus of 260,416 loosely coupled sentences and database rows. It must be noted that the semantic information in the MIMIC III database is different from the I2B2 one. For instance, there is no reason or frequency in the MIMIC III records. However, such kind of information is often present in the textual part of MIMIC III. Thus, it can be concluded that this corpus is relevant for the I2B2 extraction task. 

\subsubsection{Preprocessing and Final Datasets Distribution}

Since the seq2seq approach works at the sentence level, we extracted every medication sentence from the raw text of the MIMIC-III and i2b2 datasets using the \textit{ClarityNLP} toolkit. The tokenization was based on the \textit{Spacy} library. The language model was initialized with BERT pre-trained embeddings and then, we applied the sentence segmentation specialized for clinical documents to obtain a sentence-level segmentation and extracted the annotations in a seq2seq format as exemplified Figure~\ref{fig:annot}.

To deal with out-of-vocabulary words (OOV), BPE (Byte Pair Encoding) codes \cite{sennrich2015neural} 
were learned from the MIMIC-III and I2B2 (test set excluded) text corpora.
The final distribution of the corpus is presented Table~\ref{tab:dataDistributions}. 
In our study, we used the official test set of 251 documents (4411 sentences extracted) annotated by the community to evaluate all the models. The train set was composed of 90\% of the I2B2 unannotated documents. In the supervised learning approach, we used the freely-available MedExtractor system \cite{xu2010medex} which gave the second-best overall f-measure in the I2B2 challenge~\cite{uzuner2010community} to automatically annotate them. In the semi-supervised approach, we used a subset of the unannotated documents. 
The development set was composed of the 10 documents annotated by experts plus the remaining 10\% of the unannotated documents that were automatically annotated. Regarding MIMIC, it was only used in a semi-supervised setting. Overall, the vocabulary size of MIMIC was 43k words while I2B2 was 18k words.

\subsection{Supervised Methods}

Following the recent deep learning methods applied on the I2B2 medication extraction task, we trained the initial encoder-decoder models using simple bi-directional LSTM models with attention~\cite{luong2015effective}. This model is able to learn short and long dependencies in the input and can be trained on a reasonable amount of data. It is also surprisingly effective. We also included CNN models, since there are able to capture hierarchical relations between words and are quite efficient to train. We implemented the convolutional seq2seq model (conv-s2s) of Gehring \emph{et al.} ~\cite{gehring2017convolutional}. Finally, we included a transformer model \cite{vaswani2017attention} which are the current groundbreaking models. 

For the pre-trained word embeddings models, we explored a BERT based model \cite{devlin2018bert}. Furthermore, recently Zhu \emph{et al.}\cite{zhu2020incorporating} proposed a new algorithm for neural machine translation in which they exploit the BERT embeddings by extracting representations for an input sequence, and then fusing with each layer of the encoder and decoder through the attention mechanism. We called this model \textit{bert-fused transformer}. Finally, since in many bioNLP tasks transformer-based embeddings models pretrained on clinical data have established new baselines \cite{lee2020biobert}, we included Biobert \cite{lee2020biobert}, clinical-bert \cite{alsentzer2019publicly} and BlueBert \cite{peng2019transfer}.

\subsection{Semi-Supervised Approach}\label{sec:semiSupApproach}

For the semi-supervised learning we used the approach of Qader \emph{et al.} \cite{qader2019semi}. The approach considers two encoder-decoder models : one to extract semantics from text -- called the Natural Language Understanding (NLU) model -- and one to generate text from a semantic input -- called the natural Language Generation (NLG) model. The approach considers three sets: a paired set of texts with their annotation, a unpaired set of texts (alone) and a unpaired set of semantics annotation (alone). The paired dataset is used to learn in a supervised manner both the NLU and NLG models. The unpaired sets are used by the two modules together. The text (resp. semantic) input is fed to the NLU (resp. NLG) models which outputs a semantic representation (res. a text) which is in turn send to the NLG (resp. NLU) which outputs a text (resp. a semantic representation). The difference between the input and output  texts (resp. semantic) is used as a loss to optimize the two modules jointly. In this way, data that is not \emph{paired} with annotation can be used for learning using this `reconstruction' objective. 

\begin{table}[!bt]
\centering
{\footnotesize
\begin{tabular}{|c|c|c|c|c|c|c|c|}
\hline
\textbf{Model} & \textbf{F1} & \textbf{m} & \textbf{do} & \textbf{mo} & \textbf{f} & \textbf{du} & \textbf{r} \\ \hline
LSTM$^{*}$ & 0.78 & \textbf{0.94} & \textbf{0.92} & 0.93 & 0.89 & \textbf{0.49} & 0.50 \\ \hline
LSTM(bpe)$^{*}$ & 0.75 & 0.88 & 0.90 & 0.91 & 0.88 & 0.46 & 0.46 \\ \hline
\begin{tabular}[c]{@{}c@{}}conv-s2s\\ (bpe)$^{\dagger}$\end{tabular} & 0.68 & 0.87 & 0.83 & 0.84 & 0.76 & 0.38 & 0.41 \\ \hline
\begin{tabular}[c]{@{}c@{}}transformer\\ (bpe)$^{\dagger}$\end{tabular} & 0.75 & 0.92 & 0.88 & 0.89 & 0.84 & 0.47 & 0.50 \\ \hline\hline
\textbf{\begin{tabular}[c]{@{}c@{}}Pre-trained\\ Model\end{tabular}} & \textbf{F1} & \textbf{m} & \textbf{do} & \textbf{mo} & \textbf{f} & \textbf{du} & \textbf{r} \\ \hline
bert-base$^{\dagger}$ & 0.63 & 0.85 & 0.85 & 0.82 & 0.83 & 0.28 & 0.17 \\ \hline
\begin{tabular}[c]{@{}c@{}}bert-fused\\ -transformer$^{\dagger}$\end{tabular} & 0.74 & 0.90 & 0.87 & 0.89 & 0.83 & 0.47 & 0.50 \\ \hline
\begin{tabular}[c]{@{}c@{}}clinical-bert$^{\dagger\dagger}$\\ base\end{tabular} & 0.75 & 0.82 & 0.76 & 0.75 & 0.76 & 0.33 & 0.45 \\ \hline
biobert-base$^{\dagger\dagger}$ & 0.75 & 0.82 & 0.76 & 0.75 & 0.76 & 0.30 & 0.44 \\ \hline
\begin{tabular}[c]{@{}c@{}}bluebert-\\ base\end{tabular}$^{\dagger\dagger}$ & \textbf{0.88} & 0.92 & 0.88 & \textbf{0.95} & \textbf{0.91} & 0.46 & \textbf{0.61} \\ \hline
\end{tabular}
}
\vspace{0.1cm}
\caption{F-Measure of different models on the i2b2 medication extraction test data. ($^{*}$:seq2seq-py library , $^{\dagger}$:fairseq library , $^{\dagger\dagger}$:  https://github.com/ThilinaRajapakse/simpletransformers)}
\vspace{-0.5cm}
\label{tab:resultsSupervised}
\end{table}

As NLU and NLG models are jointly learned, the losses of the NLG and NLU models for both paired and unpaired models could be denoted respectively as $L\textsuperscript{nlg}_{paired}$ , $L\textsuperscript{nlu}_{paired}$ , $L\textsuperscript{nlg}_{unpaired}$ and $L\textsuperscript{nlu}_{unpaired}$. These four losses are mixed together to perform the joint learning $ L = \alpha~L\textsuperscript{nlg}_{paired} + \beta~L\textsuperscript{nlu}_{paired} + \gamma~ L\textsuperscript{nlg}_{unpaired} + \delta~ L\textsuperscript{nlu}_{unpaired}$ where 
$\alpha$ , $\beta$ , $\gamma$ and $\delta$ $\in [0,1]$ are fine tuned empirically.

\section{Experiments and Results}{\label{sec:parameters}}
All the experiments were performed with two open-source seq2seq libraries for the experiments \textit{seq2seq-py} from~\cite{qader2020seq2seqpy} and  fairseq library from~\cite{ott2019fairseq}.
The vocabulary size of the training corpus was around 18k tokens without BPE and 10k with BPE. Seq2Seq-py configurations used negative log-likelihood (NLL) loss, with Adam optimizer. Learning rate was set to $0.001$ with 2 bi-directional encoder-decoder layers, hidden size of $128$ and $500$ as embedding dimension. The dropout was set to $0.2$ and gradients clipping was set to $2.0$. For the fairseq experiments, LSTM architecture used cross entropy loss and nesterov accelerated gradient (NAG) as optimizer. Learning rate was fixed to $0.25$ with 4 bi-directional encoder-decoder layers. Other hyper-parameters were kept the same according to the registered model configurations of fairseq library. We kept the name of the registered architectures of fairseq for reproducibility. 
The training continued up to 70 epochs for each model and the best model was chosen according to the validation loss.

Table \ref{tab:resultsSupervised} provides the overall macro average F-measure as well as those of slot labels for all models on the I2B2 test set (see Figure~\ref{fig:annot} for the meaning of each slot label). The upper part of the table contains the results of the standard supervised models: LSTM, CNN and Transformer. To deal with OOV and vocabulary size, most of these models has been tried with BPE leading to four models (more have been evaluated but were not reported due to lack of space). 

It is clear that the simple LSTM with attention is difficult to beat since it got the highest F-measure for all slots. This might be due to the narrow domain and the lack of training data. For the same reason, using BPE does not bring any improvement. CNN and Transformer in NLP has been applied to domains with large training datasets. In this particular low-resource task, they failed to be efficient.  

Use of pre-trained embeddings leads to diverse performances. \texttt{bert-base-uncased} did not succeed in specializing enough while \texttt{bert-fused} only reached comparable performances with standard supervised methods. Again, the lack of data might explain these low performance of the general purpose pre-trained models. However, Blue Bert (\texttt{bluebert-base-uncased}) which has been specifically pre-trained on medical documents (PubMed and MIMIC) reached the best F-measure (88\%) and was particularly performing for the labels frequency (f) and reason (r) which are known to be particularly difficult.

\begin{table}[htb]
\centering
{\tiny
\begin{tabular}{|c|c|c|c|c|c|c|c|c|c|c|c|c|c|}
\hline
\textbf{corpus} & \textbf{$\alpha$} & \textbf{$\beta$} & \textbf{$\gamma$} & \textbf{$\delta$} & \textbf{\begin{tabular}[c]{@{}c@{}}F1\end{tabular}} &  \textbf{m} & \textbf{do} & \textbf{mo} & \textbf{f} & \textbf{du} & \textbf{r} \\ \hline
mimic-bpe & 1          & 0.1        & 1          & 0.1        & 0.55    & 0.83       & 0.77        & 0.79        & 0.75       & 0.37        & 0.39       \\ \hline
mimic-no-bpe & 1          & 0.1        & 1          & 0.1        & 0.64  & \textbf{0.92}       & 0.88        & 0.90        & 0.85       & \textbf{0.47}        & \textbf{0.46}       \\ \hline \hline
i2b2-bpe & 1          & 0.1        & 1          & 0.1        & 0.73  & 0.90       & 0.88        & 0.89        & 0.87       & 0.46        & 0.42       \\
\hline
i2b2-no-bpe & 1          & 0.1        & 1          & 0.1        & \textbf{0.74}  & 0.91       & \textbf{0.89}        & \textbf{0.91}        & \textbf{0.87}       & 0.43        & 0.44       \\ \hline

\end{tabular}
}
\caption{F-Measure of the semi-supervised models on the I2B2 test set using the simple LSTM model architecture.}

\vspace{-0.5cm}

\label{tab:semiSupResults}
\end{table}

Regarding the semi-supervised experiments, they have been based on the best simple LSTM from the supervised models. The results are presented table~\ref{tab:semiSupResults}. The values $\alpha$, $\beta$, $\gamma$ and $\delta$ have been fine tuned empirically. 
When MIMIC is used as unpaired data, the overall results are disappointing. This is due to the fact that MIMIC and I2B2 are still too divergent. 
When the unannotated I2B2 dataset is used as unpaired data instead of MIMIC, the performance increased. This is due to a good match between the training data and the test data. However, the performance does not reach the supervised one. Thus, it seems that for the task, using a pre-annotator like MedExtractor is more efficient than the semi-supervised strategy. However, for languages were such an extractor does not exist, the semi-supervised represents a good alternative.

\begin{table}[!b]
\centering
{\footnotesize
\begin{tabular}{|c|c|c|c|c|c|c|c|}
\hline
\textbf{System} & \textbf{F1} & \textbf{m} & \textbf{do} & \textbf{mo} & \textbf{f} & \textbf{du} & \textbf{r} \\ \hline
Guzman et al.\cite{guzman2020assessment} & 0.76 & 0.78 & 0.81 & 0.78 & 0.82 & 0.19 & - \\ \hline
Tao et al.\cite{tao2018fable} & 0.87 & 0.93 & \textbf{0.94} & \textbf{0.95} & \textbf{0.94} & \textbf{0.68} & 0.48 \\ \hline\hline
LSTM & 0.78 & \textbf{0.94} & 0.92 & 0.93 & 0.89 & 0.49 & 0.50 \\ \hline
bluebert-base & \textbf{0.88} & 0.92 & 0.88 & 0.95 & 0.91 & 0.46 & \textbf{0.61} \\ \hline
\end{tabular}
}
\caption{F-Measure of our two best models and the two the state-of-the-art models on the I2B2 test set.}\label{tab:OURvsSOTA}
\end{table}

Table \ref{tab:OURvsSOTA} summarizes the best results and compare them with the current state-of-the-art. We can see that a simple bi-LSTM model gives competitive results. Our BlueBert based pre-trained model beats the Tao \emph{et al.}~\cite{tao2018fable} model by a short margin. However, it is important to note that Tao \emph{et al.}~\cite{tao2018fable}  used a larger set of human annotated training data whereas our approach only used automatically annotated ones. Furthermore, the BlueBert model shows great capability to extract the reason slot which has been reported as the most difficult to treat in the I2B2 challenge.

\section{Discussion and future work}

This paper presents a comprehensive evaluation of the state-of-the-art seq2seq models with and without pre-trained embeddings in a supervised and semi-supervised setting. The experiments show that even with limited training data, supervised seq2seq models seems to get high slot-label prediction performance for the medical extraction task. The impact of a pre-trained model on medical documents (here BlueBert) has proven to be particularly effective in handling the lexical rich slots such as the reason and frequency concepts which are among the hardest to extract in the I2B2 2009 dataset~\cite{uzuner2010community}. The interest of such pre-trained models are in-line with other recent research in BioNLP. For the supervised learning experiments our BlueBert model reach a f-measure of 88\% in line with the current state-of-the-art method~\cite{tao2018fable} (f-measure=87\%) which used more annotated data and a complex semi-supervised pipeline. For the semi-supervised experiments, the results did not reach the state-of-the-art but the findings suggest that the unsupervised data were either too different from the test set or too small. 
Nevertheless, in case of low-resources settings, the approach could provide reasonable performances.

Future work includes how to better select and filter the unpaired datasets so that it is less noisy, closer to the target dataset and contains the most difficult cases (e.g., reason). 

Furthermore, we plan to extend this work to non-English data which are by far less resourced  \cite{Neveol2018} and thus would benefit more from an unsupervised setting. 
Finally, an ongoing future work is using this approach to develop medical prescription recognition from spoken utterances~\cite{kocabiyikoglu2019towards}. This would have many applications, from harvesting large amount of spoken clinical data, to building medical assistants on smartphones. These applications would increase traceability in health care centers and could reduce the number of medication errors.

\bibliographystyle{IEEEtran}

\bibliography{biblio}

\end{document}